\documentclass[10pt,twocolumn,letterpaper]{article}

\usepackage{iccv}
\usepackage{times}
\usepackage{epsfig}
\usepackage{graphicx}
\usepackage{amsmath}
\usepackage{amssymb}
\usepackage{bbding}

\usepackage[numbers,sort,compress]{natbib}
\usepackage{soul}
\usepackage[dvipsnames]{xcolor}
\usepackage{xspace}
\usepackage[normalem]{ulem}

\usepackage{tabularx,ragged2e,booktabs}
\usepackage{multirow}

\usepackage{color, colortbl}
\definecolor{Gray}{gray}{0.9}

\usepackage{enumitem}

\usepackage[heightadjust=all,valign=c]{floatrow}
\newfloatcommand{capbtabbox}{table}[][\FBwidth]

\usepackage[outline]{contour}
\usepackage{mathtools}
\usepackage{bbm}
\usepackage{pifont}
\newcommand{\cmark}{\ding{51}}
\usepackage{eso-pic}

\usepackage{sg-macros} 

\makeatletter
\renewcommand{\paragraph}{%
  \@startsection{paragraph}{4}%
  {\z@}{1.05ex \@plus 1ex \@minus .2ex}{-1em}%
  {\normalfont\normalsize\bfseries}%
}
\makeatother

\usepackage[breaklinks=true,bookmarks=false]{hyperref}

\iccvfinalcopy 

\def\iccvPaperID{****} 

\begin{document}

\title{A Self-Training Framework Based on Multi-Scale
Attention Fusion for Weakly Supervised Semantic
Segmentation\\
}

\author{Guoqing Yang \quad 
Chuang Zhu\footnotemark[1] \quad
Yu Zhang\\ 
Beijing University of Posts and Telecommunications, China\\
\texttt{\{yangguoqing, czhu, zhangyu\_03\}@bupt.edu.cn}
}

\maketitle

\begin{abstract}
Weakly supervised semantic segmentation (WSSS) based on image-level labels is challenging since it is hard to obtain complete semantic regions.
To address this issue,
we propose a self-training method that utilizes fused multi-scale class-aware attention maps.
Our observation is that attention maps of different scales contain rich complementary information, especially for large and small objects.
Therefore, 
we collect information from attention maps of different scales and obtain multi-scale attention maps. 
We then apply denoising and reactivation strategies to enhance the potential regions and reduce noisy areas. 
Finally, we use the refined attention maps to retrain the network. 
Experiments showthat our method enables the model to extract rich semantic information from multi-scale images and achieves 72.4\% mIou scores on both the PASCAL VOC 2012 validation and test sets. 
The code is available at \href{https://bupt-ai-cz.github.io/SMAF}
{https://bupt-ai-cz.github.io/SMAF.}


\end{abstract}
\footnotetext[1]{coressponding author~(czhu@bupt.edu.cn).}

\section{Introduction}

As an important task in computer vision, semantic segmentation plays an important role in many fields. 
However, 
training a fully supervised semantic segmentation requires  dense annotations,
which can be laborious and time-consuming to obtain accurately. 
To address this issue, 
weakly supervised semantic segmentation~(WSSS)~is introduced, 
which only requires coarse labels such as image-level labels\cite{wei2017object,zeng2019joint,lee2021railroad}, scribbles\cite{lin2016scribblesup,xu2021scribble},
bounding boxes\cite{dai2015boxsup,lee2021bbam},
and points\cite{bearman2016s,chen2021seminar}.
Among these approaches, 
WSSS based on image-level labels has attracted the most attention for its low cost.
Therefore this paper focuses on the WSSS based on image-level labels.

\begin{figure}[htbp] 
    \centering
    \includegraphics[width=0.95\columnwidth]{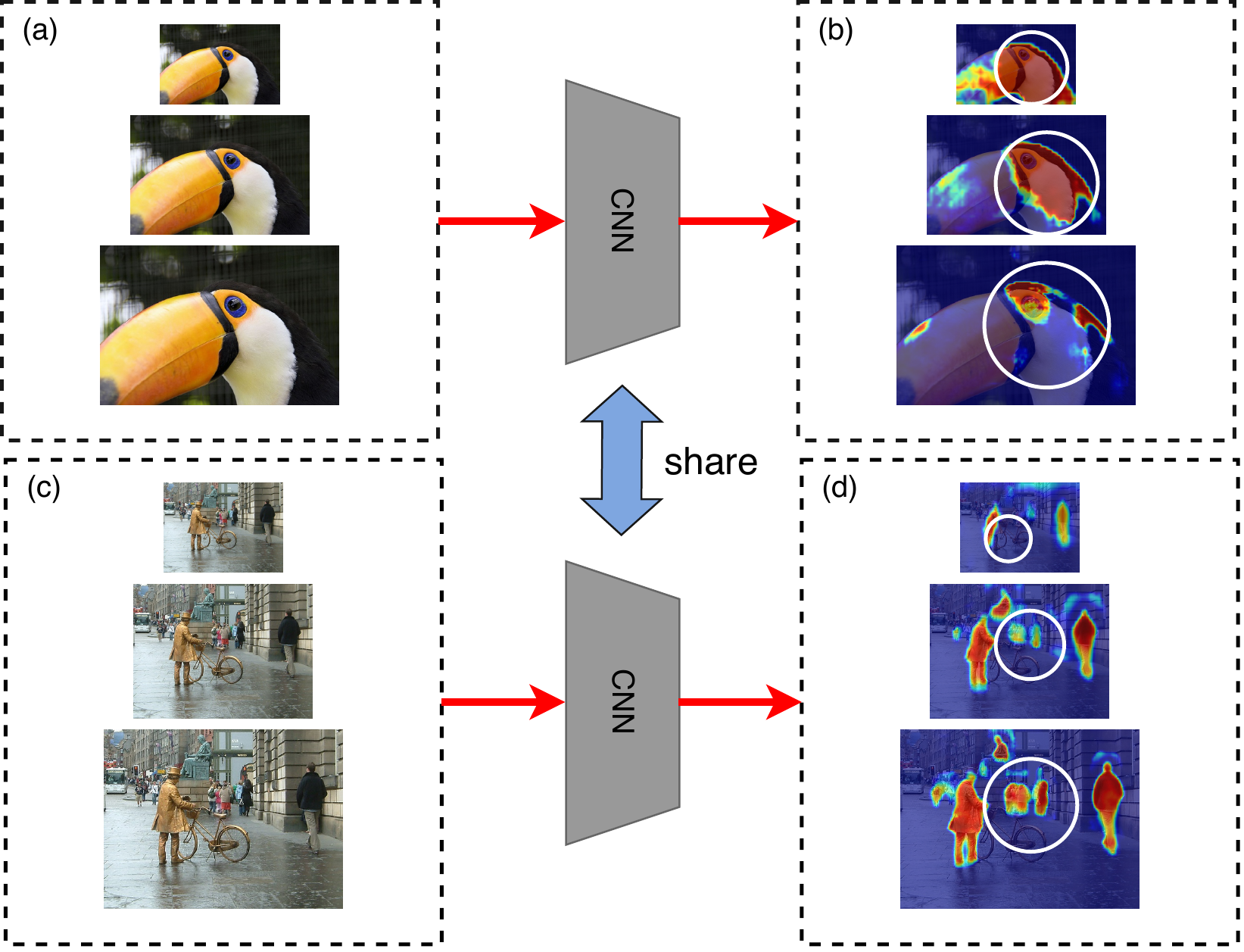}
    \caption{The motivation of our  proposed method.
    We visualize the attention maps generated by input images at different scales.
    (a) and (b): large objects~(a bird covering most of the image area)~and their corresponding attention maps;
    (c) and (d): small objects~(people in the distance on the street, covering a small area in the image)~and their corresponding attention maps.
    }
    \label{motivation}
\end{figure}

For most existing methods,
Class Activation Mapping (CAM)\cite{zhou2016learning} is adopted to provide initial location cues and used as pseudo segmentation labels for training the semantic segmentation model.
However, 
class-aware attention maps,
known as CAMs,
tend to focus on the most discriminative regions, 
which have a gap with the dense annotation required for semantic segmentation.
Many strategies have been proposed to narrow this gap,
such as region erasing and growing\cite{wei2017object, huang2018weakly,shimoda2019self},
using additional supervision information\cite{yao2020saliency,chang2020weakly,lee2022weakly},
and self-supervised learning\cite{zhou2022regional,du2022weakly}.
Despite their good performance, 
there is still untapped potential to further improve the WSSS model.

Previous studies,
such as\cite{wang2020self} have demonstrated that the responses of a WSSS model can differ when presented with images of different scales.
We further investigate this phenomenon and observe that these differences exhibit a certain level of complementarity that is related to the size of objects in the image.
As illustrated in Fig.~\ref{motivation}, attention maps generated from enlarged images tend to miss overall semantic information for large objects, whereas those from reduced images can capture it better. Conversely, attention maps generated from reduced images may lose some targets for small objects, but those from enlarged images can help recover them. Hence, it is promising to collect information from attention maps at different scales for training the model's single-scale responses.

To this end, 
we propose a self-training framework that utilizes multi-scale attention maps to improve the performance of the model.
Specifically, we first generate attention maps at different scales for a given image and then fuse them using a fusion strategy to produce initial multi-scale attention maps.
Both enlarged and reduced transformations are required for this purpose. However, the initial multi-scale attention maps often contain noisy and under-activated regions. To refine them, we apply denoising and reactivation strategies. We then use the refined multi-scale attention maps to supervise the network's response to single-scale images. One advantage of our framework is that by incorporating information from different scales, it can help the model overcome bias towards single-scale images and capture more complete semantic regions.

In common practice\cite{wang2020self,lee2021railroad}, 
the multi-scale method is often directly used in the inference stage to generate pseudo segmentation labels.
In contrast,
we refine the multi-scale attention maps by using denoising and reactivation strategies and then use them to supervise the model’s response to single-scale images.
As a result, 
as shown in Fig.~\ref{attention}, 
our model can capture more target regions.
Furthermore, 
our framework is flexible and can apply to any WSSS model.

In summary, our contributions are as follows:
\begin{itemize}
    \item We investigate the response of different image scales in the WSSS model and find that large and small objects exhibit complementary behavior when images are resized to different scales.
    \item We propose a self-training method that utilizes fused multi-scale attention maps to enhance the model's ability for mining semantic features. 
    Specifically, we take into account the effects of image enlargement and reduction and employ denoising and reactivation strategies to refine the multi-scale attention maps.
    \item Our method significantly outperforms the baseline model, achieving 72.4\% mIou scores both on the PASCAL VOC 2012 val and test sets.
\end{itemize}

\begin{figure*}[htbp]
    \centering
    \includegraphics[width=170mm]{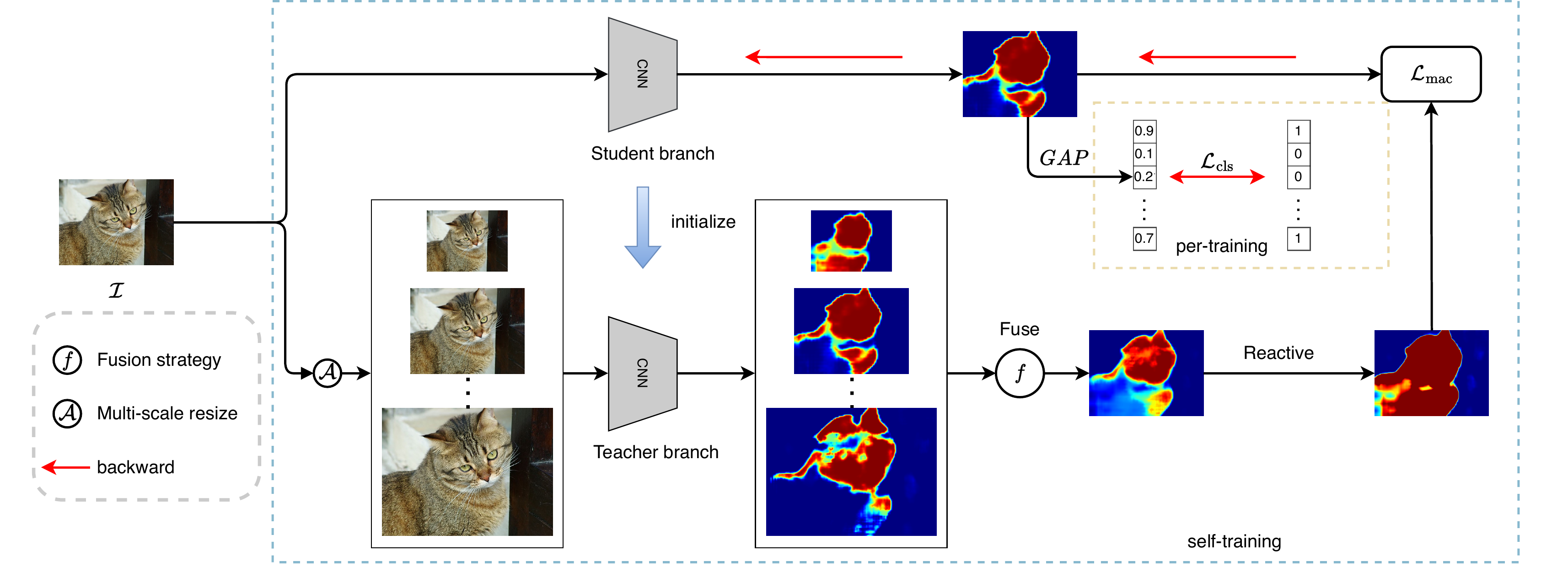}
    \caption{Overview of our proposed method.
    We first pre-train the student branch using an existing WSSS method and initialize the teacher branch.
    The teacher branch is responsible for generating fused multi-scale attention maps, which are then refined by denoising and reactivation strategies. Finally, the refined multi-scale attention maps are used to train the student branch.}
    \label{main struct}
\end{figure*}

\section{Related Work}
\paragraph{Image-level WSSS}
has received extensive research due to its high efficiency.
The two-stage image-level WSSS follows the pipeline that generates pseudo segmentation labels and trains a fully supervised segmentation network.
Recent WSSS methods relay on CAMs\cite{zhou2016learning} to extract location information from images and image-level labels.
However,
CAMs only capture the most discriminative regions of objects.
The intrinsic reason for this phenomenon is the gap between classification and segmentation tasks.
Only crucial information for classification can flow to the classification layer\cite{lee2021reducing}.
Consequently, the pseudo segmentation labels obtained from CAMs are often inaccurate.

To address this issue,
some studies enforce networks to pay more attention to non-discriminative regions using discriminative region erasing\cite{wei2017object,zhang2018adversarial},
region growing\cite{huang2018weakly,shimoda2019self}.
Some studies have introduced additional supervision information, such as saliency maps\cite{yao2020saliency,lee2021railroad,zeng2019joint},
cross images\cite{sun2020mining},
sub-categories\cite{chang2020weakly} and out-of-distribution data\cite{lee2022weakly}.
Self-supervised learning has also been employed in some works to extract information, 
such as SEAM\cite{wang2020self},which proposes consistency regularization on predicted CAMs from various transformed images.
RCA\cite{zhou2022regional} and PPC\cite{du2022weakly} leverage contrastive learning to ensure that pixels sharing the same labels have similar representations in the feature space, and vice versa. Recently, with the emergence of Transformer, some studies\cite{rossetti2022max,xu2022multi} have attempted to replace CNN with Transformer and achieved promising results.

\section{Proposed Method}
The entire framework is illustrated in Fig.~\ref{main struct}.
In this section, we first introduce the generation of class-aware attention maps.
Then we describe the multi-scale attention fusion strategy and reactive strategy.
Subsequently, 
we use the fused multi-scale attention maps to train the model.
The overall loss function is formulated as: 
\begin{equation}
  \mathcal{L}_{total}=\mathcal{L}_{cls}+\alpha\mathcal{L}_{mac},
\end{equation}
where $\mathcal{L}_{total}$ denotes the overall loss, $\mathcal{L}_{cls}$ is the classification loss, and $\mathcal{L}_{mac}$ is the multi-scale attention consistency loss. The hyperparameter $\alpha$ is used to balance the two components of the loss function.

\subsection{Class-awared Attention Maps}

Given image $\mathcal{I}$ and image-level labels~$y\in\mathbb{R}^{K}$, where $K$ is the number of categories present in the dataset.
We can obtain the class-aware attention maps from the last convolutional layer of the network:
\begin{equation}
    \mathcal{M}=\mathbf{ReLU}(\boldsymbol{f}(\mathcal{I})),
\end{equation}
where $\mathcal{M}$ is the class-aware attention maps with the spatial size of $K\times H\times W$,
and $\boldsymbol{f}(\cdot)$ is the backbone.
After the $\mathbf{ReLU(\cdot)}$ activation function, 
the attention maps are normalized to ensure that their scores are distributed within the range of [0, 1].
The last convolutional layer is followed by a global average pooling~(GAP)~layer to obtain the image-level prediction~$\hat{y}\in\mathbb{R}^{K}$,
which is used to train a classifier using the cross-entropy loss function:
\begin{equation}
  \mathcal{L}_{cls}=\frac{1}{K}\sum\limits_{k=0}^{K-1}{y}^{k}log\sigma(\hat{y}^{k})+(1-{y}^{k})log(1-\sigma(\hat{y}^{k})),
\end{equation}
where $\sigma(\cdot)$ is the sigmoid function.
Once the classifier is well trained, we can utilize $\mathcal{M}$ to generate pseudo segmentation labels:
\begin{equation}
  \mathbf{P}=argmax(\mathcal{M}),
\end{equation}
where $\mathbf{P}$ denotes the generated pseudo segmentation labels with the spatial size of $H\times W$.

\subsection{Multi-scale Attentions Fusion Strategy}

Fig.~\ref{main struct} illustrates the overall process of our approach. Prior to self-training, we pre-train the student branch using existing WSSS techniques with image-level labels. We then initialize the teacher branch with the pre-trained model, which has a preliminary segmentation ability. For this purpose, we adopt EPS\cite{lee2021railroad} for its performance and conciseness.

In the following, 
we describe how we fuse the different scales of attention maps.
Firstly,
we resize the original image~$\mathcal{I}$ to different scales,
denoted as $\mathcal{I}^{'}=\{\mathcal{I}_{s},\mathcal{I}_{o},\mathcal{I}_{l}\}$,
where $\mathcal{I}_{s},\mathcal{I}_{o},\mathcal{I}_{l}$ represent the small-scale, original, and large-scale images,
respectively.
We then obtain their corresponding class-aware attention maps $\mathcal{M}^{'}=\{\mathcal{M}_{s},\mathcal{M}_{o},\mathcal{M}_{l}\}$.
It is worth noting that we consider both large-scale and small-scale transformations to take full advantage of the complementary information.
Next, 
we fuse $\mathcal{M}^{'}$ to integrate the complementary information.
In this study,
we propose a fusion strategy that involves averaging attention maps, 
which is commonly used in WSSS during the inference stage:
Specifically, 
the fused attention map $ \mathcal{F}^{k}$ for the  $k$-th channel is calculated as follows:
\begin{equation}
  \mathcal{F}^{k}=\frac{\mathcal{M}_{k}}{max(\mathcal{M}_{k})}, k\in K,
\end{equation}
where $ \mathcal{F}^{k}$ denotes the $k$-th channel of the fused attention map.
The calculation of $ \mathcal{M}_{k}$ is performed as follows:
\begin{equation}
 \mathcal{M}_{k}=\mathcal{M}_{s}^{k}+\mathcal{M}_{o}^{k}+\mathcal{M}_{l}^{k},
\end{equation}
where $\mathcal{M}_{s}^{k}$,$\mathcal{M}_{o}^{k}$ and $\mathcal{M}_{l}^{k}$ represent the $k$-th channel of the attention maps for the small-scale, original, and large-scale images, respectively.
As the attention maps can vary in size across different scales,
we resize them to the same size as $\mathcal{M}_{o}$ before adding them together.
To restrict the range of the attention scores to [0, 1], we normalize the $k$-th channel by the maximum value of $\mathcal{M}_{k}$, 
which is denoted as $max(\mathcal{M}_{k})$. 

The $\mathcal{F}$ contains complementary information from different scales of attention maps.
Compared to the single-scale attention maps $\mathcal{M}_{o}$ from the student branch, 
$\mathcal{F}$ can capture more target regions.
To measure the difference between $\mathcal{F}$ and $\mathcal{M}_{o}$, we use the multi-scale attention consistency loss $\mathcal{L}_{mac}$, defined as follows:

\begin{equation}
  \mathcal{L}_{mac}=\frac{1}{K}\sum\limits_{k}\Vert \mathcal{F}^{k}-\mathcal{M}_{o}^{k} \Vert^{2}, k\in K.
\end{equation}
Here,
$\mathcal{F}^{k}$ and $\mathcal{M}_{o}^{k}$ represent the $k$-th channel of the fused attention maps and the output of the student branch, respectively. 
The $\Vert \cdot \Vert^{2}$ is given by
$\frac{1}{H\times W}\sum\limits_{i}\sum\limits_{j}(\mathcal{F}^{k}_{i,j}-\mathcal{M}^{k}_{o~i,j})^{2}$, 
where $(i,j)$ represents the coordinates of the pixel and ${H}$ and $W$ are the height and width of the attention maps.

\begin{figure*}[htbp]
    \centering
    \includegraphics[width=170mm]{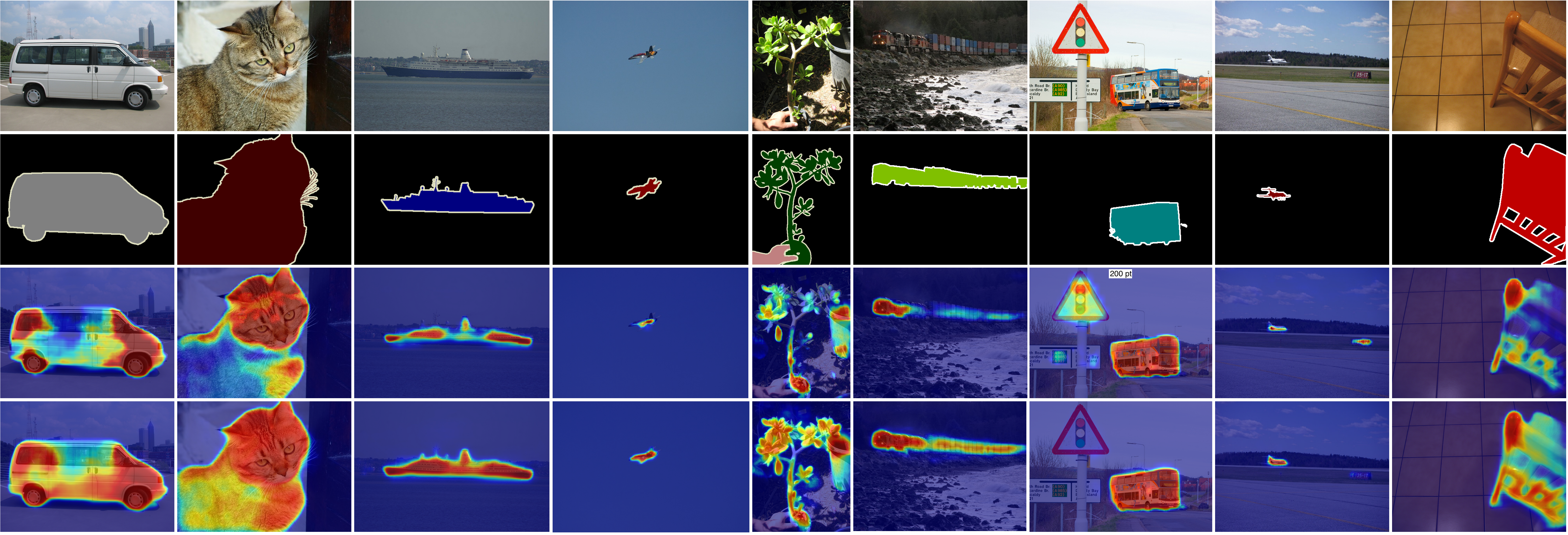}
    \caption{Visual comparison of attention maps quality.
    From top to bottom:
    original image,
    ground truth,
    attention maps generated by EPS\cite{lee2021railroad},
    and attention maps generated by our method.
    }
    \label{attention}
\end{figure*}

\begin{figure}[htbp] 
    \centering
    \includegraphics[width=0.95\columnwidth]{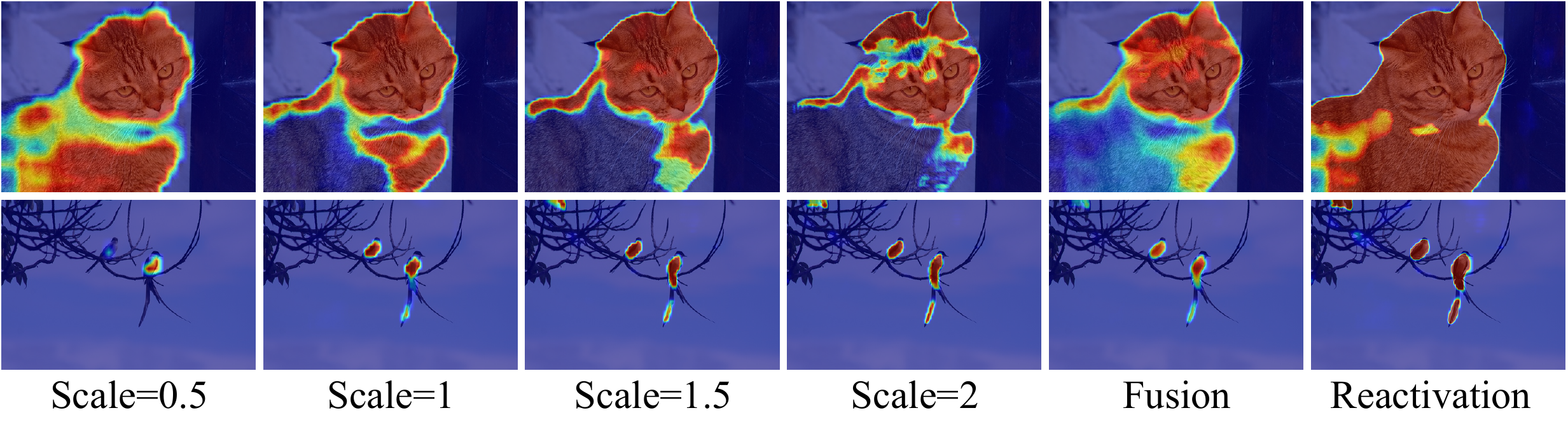}
    \caption{Visualization of different attention maps.
    }
    \label{compare_reac}
\end{figure}

\subsection{Denoising and Reactivation Strategies}

The $\mathcal{F}$ still has several flaws, including noisy and under-activated areas.
we propose to incorporate image-level labels for inter-channel denoising. 
Specifically,
if class $k$ is not present in $y$, 
we set the values of the corresponding channel in $\mathcal{F}$ to 0.

Furthermore, as shown in Fig.~\ref{compare_reac},
~$\mathcal{F}$  can capture more complete regions than $\mathcal{M}_{o}$. 
However,
we also observe that some regions may be under-activated in $\mathcal{F}$ if they are only activated in a single-scale attention map, which can be detrimental to the training of the student branch. 
To address this issue, 
we introduce a reactivation strategy to refine $\mathcal{F}$. 
Specifically,
we first set the values in background channel to threshold $thr$ and then apply the following formula to reactivate these areas:
\begin{equation}
  \mathcal{F}_{i}^{'k}=\frac{\mathcal{F}_{i}^{k}}{\mathop{\max}\limits_{k}(\mathcal{F}_{i}^{k})}, k\in K,
\end{equation}
where $\mathcal{F}_{i}^{'k}$ is the value of reactivated attention maps for pixel $i$,
and $\mathcal{F}^{k}$ is set to 0 for $k$ not present in the image-level labels $y$.
Finally, 
we define the attention consistency loss as follows:
\begin{equation}
  \mathcal{L}_{mac}=\frac{1}{K}\sum\limits_{k}\Vert \mathcal{F}^{'k}-\mathcal{M}_{o}^{k} \Vert^{2}, k\in K.
\end{equation}

\section{Experiments}

\subsection{Experimental Settings}

\paragraph{Dataset and Evaluated Metric}
This study is conducted on the PASCAL VOC 2012 dataset\cite{everingham2010pascal},which serves as the standard benchmark in WSSS. This dataset consists of 20 semantic categories and a background, and comprises 1,464, 1,449, and 1,456 images for the training, validation, and test sets, respectively. To enhance the training set, we use the SBD augmented training set\cite{hariharan2011semantic},as has been done in previous studies,which provides 10,582 images. The performance of our approach is evaluated using the mean intersection-over-union (mIoU)\cite{long2015fully}.

\begin{table}[tp]
\begin{center}
\caption{Segmentation performance mIoU (\%) on Pascal VOC 2012 val and test sets using DeepLab-ASPP.
*~means using saliency maps.} 
\begin{tabular}{|c|c|c|c|}
  \hline
  \textbf{Methods}       & \textbf{Backbone}       & \textbf{val}   & \textbf{test}    \\
  \hline
  PSA\cite{ahn2018learning}$_{~\textup{CVPR’18}}$    & ResNet38         &61.7      & 63.2      \\
    IRN\cite{ahn2019weakly}$_{~\textup{CVPR’19}}$   & ResNet50          &63.5      & 64.8       \\
    ICD$^{*}$\cite{fan2020learning}$_{~\textup{CVPR’20}}$     &ResNet101         & 64.1   &64.3     \\
    SEAM\cite{wang2020self}$_{~\textup{CVPR’20}}$   & ResNet38         & 64.5    &65.7    \\
    MCIS\cite{sun2020mining}$_{~\textup{ECCV’20}}$   & ResNet101      &66.2   &66.9      \\
    EDAM$^{*}$\cite{wu2021embedded}$_{~\textup{CVPR’21}}$   & ResNet101         &70.9    &70.6     \\
    AdvCAM\cite{lee2021anti}$_{~\textup{CVPR’21}}$   & ResNet101          &68.1    &68.0     \\
    SIPE\cite{chen2022self}$_{~\textup{CVPR’21}}$   & ResNet101           &68.8    &  69.7   \\
    L2G$^{*}$\cite{jiang2022l2g}$_{~\textup{CVPR’22}}$   & ResNet101      &72.1    & 71.7    \\
    \hline
    EPS$^{*}$\cite{lee2021railroad}$_{~\textup{CVPR’21}}$      & ResNet101      &70.9 &70.8   \\
    \textbf{Ours w/EPS}    &  ResNet101  &\textbf{72.4}\textcolor{red}{$_{\uparrow \textbf{1.5}}$}  &\textbf{72.4}\textcolor{red}{$_{\uparrow \textbf{1.6}}$} \\
  \hline
\end{tabular}
\label{tab:Compare_seg}
\end{center}
\end{table}

\begin{table}[tp]
\begin{center}
\caption{Evaluation (mIoU (\%)) of the initial attention maps (Seed),
refined by CRF (+CRF)~on PASCAL VOC 2012 train set.} 
\begin{tabular}{|c|c|c|}
  \hline
  \textbf{Methods}  & \textbf{Seed}  &  \textbf{+DenseCRF}
  \\
  \hline
  ICD\cite{fan2020learning}$_{~\textup{CVPR’20}}$          & 59.9             & 62.2   \\
    SEAM\cite{wang2020self}$_{~\textup{CVPR’20}}$            & 55.4             & 56.8   \\
    EDAM\cite{wu2021embedded}$_{~\textup{CVPR’21}}$           & 52.8             & 58.2   \\
    SIPE\cite{chen2022self}$_{~\textup{CVPR’22}}$     & 58.6      & 64.7        \\
    PPC w/EPS\cite{zhou2022regional}$_{~\textup{CVPR’22}}$    & 70.5      & 73.3        \\
    \hline
    EPS\cite{lee2021railroad}$_{~\textup{CVPR’21}}$            & 69.5         & 71.4        \\
    \textbf{Ours w/EPS}     & \textbf{72.0}\textcolor{red}{$_{\uparrow \textbf{2.5}}$}      &  \textbf{73.8}\textcolor{red}{$_{\uparrow \textbf{2.4}}$}  \\
  \hline
\end{tabular}
\label{tab:Compare_pm}
\end{center}
\end{table}

\paragraph{Implementation Details}

\begin{figure*}[htbp]
    \centering
    \includegraphics[width=170mm]{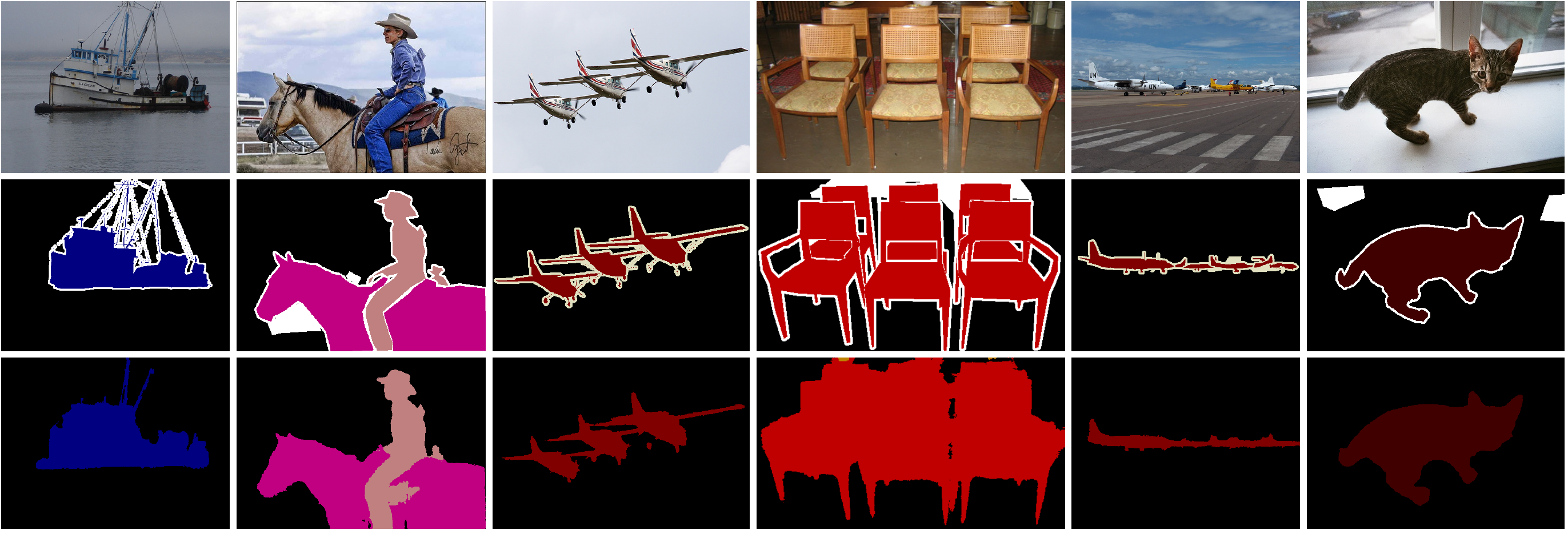}
    \caption{Qualitative segmentation results on PASCAL VOC 2012 val set.~From top to bottom: input images, ground truth,  segmentation results of our method.}
    \label{segdeeplab2}
\end{figure*}

Following the common WSSS works,
we adopt ResNet38\cite{wu2019wider} as our backbone.
Prior to self-training, 
we pre-train the student branch with EPS\cite{lee2021railroad} and initialize the teacher branch with the pre-trained model.
To augment our input images, we implement the data augmentation strategies following\cite{lee2021railroad,wang2020self} for the student branch. 
For the teacher branch, 
we resize the original images with scales of $\{0.5,1,1.5,2\}$ and apply flipping operation.
During self-training,
we employ SGD with a batch size of 8, momentum of 0.9, and weight decay of 5e-4 as the optimizer for the student branch.
We train the network for 20k iterations, with the teacher branch being frozen during this process. For hyper-parameters, we empirically set $\alpha$ and $thr$ to 100 and 0.2, respectively.

Once our model is trained, we follow the inference procedure outlined in other WSSS works to generate pseudo segmentation labels using Dense-CRF\cite{krahenbuhl2011efficient}. During inference, the student branch generates the pseudo segmentation labels, while the teacher branch is discarded. Finally, with the supervision of the pseudo segmentation labels, we train Deeplab-ASPP\cite{chen2017deeplab} using the default parameters. Standard Dense-CRF is employed as a post-processing step to refine the final segmentation results.

\subsection{Comparison with State-of-the-arts}

\subsubsection{Class-aware Attention Maps}

\begin{table}[htbp]
\begin{center}
\caption{The comparison of the impact for different attention fusion strategies.} \label{tab:Multi}
\begin{tabular}{|c|c c c c|c|}
  \hline
  \textbf{Fusion strategy} & \textbf{0.5}& \textbf{1}& \textbf{1.5}& \textbf{2}  & \textbf{mIoU(\%)}    \\
  \hline
  Small-scale       & \cmark  & \cmark  & &   &71.1         \\
    Large-scale 1    &   &\cmark  &\cmark     & & 70.5        \\
    Large-scale 2   &   &\cmark &  &\cmark     & 70.5      \\
    \hline
    \textbf{Full-scale}  &\cmark  &\cmark  &\cmark  &\cmark & \textbf{72.0}       \\
  \hline
\end{tabular}
\end{center}
\vspace{-0.5cm}
\end{table}

Table~\ref{tab:Compare_pm}
presents the mIou scores of pseudo segmentation labels obtained from PASCAL VOC 2012 train set. 
Following EPS\cite{lee2021railroad},
we directly obtain the seeds from the network, without resorting to additional post-processing operations such as random walk,
PSA\cite{ahn2018learning},
or IRN\cite{ahn2019weakly}. 
As the common practice,
we utilize Dense-CRF for refining the seeds to generate the final pseudo segmentation labels.
Notably, our approach yields an improvement of 3.4\% and 2.4\% in terms of mIou scores over EPS\cite{lee2021railroad} for seed and seed + Dense-CRF, respectively.


As shown in Fig.~\ref{attention},
our method exhibits excellent performance on both large and small objects in the image.
This result suggests that learning from the complementary information provided by fused multi-scale attention maps leads to more accurate feature expressions.


\subsubsection{Segmentation Results}

Following the common practice, 
we employ pseudo segmentation labels to train Deeplab-ASPP to make a fair comparison. Table~\ref{tab:Compare_seg} indicates that our approach improves the EPS\cite{lee2021railroad} by 1.5\% in terms of mIou score on both the val and test sets of PASCAL VOC 2012. This outcome establishes the effectiveness of our method in enhancing the performance of the initial WSSS model without the need for external data. Fig.~\ref{segdeeplab2} depicts some segmentation results obtained from the PASCAL VOC 2012 val set.

\subsection{Ablation Studies}

To demonstrate the effectiveness of each component, 
we conduct ablation studies on the PASCAL VOC 2012 train set.

\begin{table}[htp]
\begin{center}
\caption{Experimental comparison between using single-scale and multi-scale attention maps as self-training supervision.} 
\begin{tabular}{|c|c|c|}
  \hline
  \textbf{Method}      & \textbf{Scale}  & \textbf{mIoU(\%)}    \\
  \hline
    \multirow{4}{*}{Single-scale} & 0.5                   &69.2              \\
                                    & 1                  &    70.7           \\
                                    & 1.5              &   67.7         \\
                                    & 2          &   63.0      \\
    \hline
    \textbf{Multi-scale}                  & \textbf{All}   &  \textbf{72.0}        \\
  \hline
\end{tabular}
\label{tab:Single}
\end{center}
\end{table}

\subsubsection{Single-scale vs. Multi-scale}

To begin with, we evaluate the benefits of using multi-scale attention maps over single-scale attention maps. 
Specifically, 
we employ attention maps generated from different scales of images, 
along with our fused multi-scale attention maps, 
to train the student branch. 
The selected scale factors are $\left\{0.5,1,1.5,2\right\}$ respectively. 
The corresponding results are reported in Table~\ref{tab:Single}.

Notably, compared to the single-scale approach, the model trained with the fused attention maps for self-training achieves performance improvements of $\left\{2.8\%,1.3\%,4.3\%,9\%\right\}$ for the respective scales. This finding suggests that the attention maps from different scales only provide partial information, and simply relying on single-scale attention maps could be detrimental to the network's performance. Fig.~\ref{compare_reac} further illustrates the difference between the approaches.

\subsubsection{Multi-scale Fusion Strategy}

We also investigate the impact of various fusion strategies, namely, small-scale, large-scale, and full-scale. 
It is important to note that the full-scale fusion strategy encompasses both enlarged and reduced transformations. 
The results are summarized in Table~\ref{tab:Multi}, where the full-scale fusion strategy yields mIoU scores that are 0.9\% and 1.5\% higher than those obtained by the small-scale and large-scale fusion strategies, respectively.

\subsubsection{Attention Reactivation Strategy}
Table~\ref{tab:reactive}~presents the impact of the reactivation strategy on the PASCAL VOC 2012 training set. It is noteworthy that all experiments are conducted based on a full-scale fusion strategy. The results show that the removal of the reactivation strategy leads to a 0.9\% decrease in the mIoU score. This finding highlights the beneficial role of reactivation in self-training. Furthermore, Fig.~\ref{compare_reac} demonstrates that this strategy effectively enhances the under-activated regions in attention maps.

\section{Conclusion}
\begin{table}[htp]
\begin{center}
\caption{The comparision of the impact for reactivation strategy.} 
\begin{tabular}{|c|c|c|}
  \hline
  \textbf{variant} & \textbf{w/o Reactivation}& \textbf{w/ Reactivation}    \\
  \hline
  \textbf{mIoU(\%)}    & 71.1     &72.0         \\
  \hline
\end{tabular}
\label{tab:reactive}
\end{center}
\end{table}

In this work, 
we propose a self-training framework that employs a multi-scale attention fusion method to enhance the performance of image-level WSSS.
Our framework utilizes complementary information from different scales of attention maps to supervise the model's response to single-scale images. 
Moreover, we adopt denoising and reactivation strategies to refine the fused attention maps. 
We evaluate our proposed method extensively on the PASCAL VOC 2012 dataset and demonstrate its effectiveness in improving the performance of image-level WSSS.

\section*{Acknowledgement}
This work was supported by the National Key R\&D Program of China (2021ZD0109802) and the National Natural Science Foundation of China (81972248).

{\small
\bibliographystyle{ieee_fullname}
\bibliography{egbib}

\begin{thebibliography}{10}\itemsep=-1pt

\bibitem{ahn2019weakly}
Jiwoon Ahn, Sunghyun Cho, and Suha Kwak.
\newblock Weakly supervised learning of instance segmentation with inter-pixel
  relations.
\newblock In {\em CVPR}, pages 2209--2218, 2019.

\bibitem{ahn2018learning}
Jiwoon Ahn and Suha Kwak.
\newblock Learning pixel-level semantic affinity with image-level supervision
  for weakly supervised semantic segmentation.
\newblock In {\em CVPR}, pages 4981--4990, 2018.

\bibitem{bearman2016s}
Amy Bearman, Olga Russakovsky, Vittorio Ferrari, and Li Fei-Fei.
\newblock What’s the point: Semantic segmentation with point supervision.
\newblock In {\em ECCV}, pages 549--565. Springer, 2016.

\bibitem{chang2020weakly}
Yu-Ting Chang, Qiaosong Wang, Wei-Chih Hung, Robinson Piramuthu, Yi-Hsuan Tsai,
  and Ming-Hsuan Yang.
\newblock Weakly-supervised semantic segmentation via sub-category exploration.
\newblock In {\em CVPR}, pages 8991--9000, 2020.

\bibitem{chen2021seminar}
Hongjun Chen, Jinbao Wang, Hong~Cai Chen, Xiantong Zhen, Feng Zheng, Rongrong
  Ji, and Ling Shao.
\newblock Seminar learning for click-level weakly supervised semantic
  segmentation.
\newblock In {\em ICCV}, pages 6920--6929, 2021.

\bibitem{chen2017deeplab}
Liang-Chieh Chen, George Papandreou, Iasonas Kokkinos, Kevin Murphy, and Alan~L
  Yuille.
\newblock Deeplab: Semantic image segmentation with deep convolutional nets,
  atrous convolution, and fully connected crfs.
\newblock {\em TPAMI}, 40(4):834--848, 2017.

\bibitem{chen2022self}
Qi Chen, Lingxiao Yang, Jian-Huang Lai, and Xiaohua Xie.
\newblock Self-supervised image-specific prototype exploration for weakly
  supervised semantic segmentation.
\newblock In {\em CVPR}, pages 4288--4298, 2022.

\bibitem{dai2015boxsup}
Jifeng Dai, Kaiming He, and Jian Sun.
\newblock Boxsup: Exploiting bounding boxes to supervise convolutional networks
  for semantic segmentation.
\newblock In {\em ICCV}, pages 1635--1643, 2015.

\bibitem{du2022weakly}
Ye Du, Zehua Fu, Qingjie Liu, and Yunhong Wang.
\newblock Weakly supervised semantic segmentation by pixel-to-prototype
  contrast.
\newblock In {\em CVPR}, pages 4320--4329, 2022.

\bibitem{everingham2010pascal}
Mark Everingham, Luc Van~Gool, Christopher~KI Williams, John Winn, and Andrew
  Zisserman.
\newblock The pascal visual object classes (voc) challenge.
\newblock {\em IJCV}, 88(2):303--338, 2010.

\bibitem{fan2020learning}
Junsong Fan, Zhaoxiang Zhang, Chunfeng Song, and Tieniu Tan.
\newblock Learning integral objects with intra-class discriminator for
  weakly-supervised semantic segmentation.
\newblock In {\em CVPR}, pages 4283--4292, 2020.

\bibitem{hariharan2011semantic}
Bharath Hariharan, Pablo Arbel{\'a}ez, Lubomir Bourdev, Subhransu Maji, and
  Jitendra Malik.
\newblock Semantic contours from inverse detectors.
\newblock In {\em ICCV}, pages 991--998. IEEE, 2011.

\bibitem{huang2018weakly}
Zilong Huang, Xinggang Wang, Jiasi Wang, Wenyu Liu, and Jingdong Wang.
\newblock Weakly-supervised semantic segmentation network with deep seeded
  region growing.
\newblock In {\em CVPR}, pages 7014--7023, 2018.

\bibitem{jiang2022l2g}
Peng-Tao Jiang, Yuqi Yang, Qibin Hou, and Yunchao Wei.
\newblock L2g: A simple local-to-global knowledge transfer framework for weakly
  supervised semantic segmentation.
\newblock In {\em CVPR}, pages 16886--16896, 2022.

\bibitem{krahenbuhl2011efficient}
Philipp Kr{\"a}henb{\"u}hl and Vladlen Koltun.
\newblock Efficient inference in fully connected crfs with gaussian edge
  potentials.
\newblock {\em NIPS}, 24, 2011.

\bibitem{lee2021reducing}
Jungbeom Lee, Jooyoung Choi, Jisoo Mok, and Sungroh Yoon.
\newblock Reducing information bottleneck for weakly supervised semantic
  segmentation.
\newblock {\em NIPS}, 34:27408--27421, 2021.

\bibitem{lee2021anti}
Jungbeom Lee, Eunji Kim, and Sungroh Yoon.
\newblock Anti-adversarially manipulated attributions for weakly and
  semi-supervised semantic segmentation.
\newblock In {\em CVPR}, pages 4071--4080, 2021.

\bibitem{lee2022weakly}
Jungbeom Lee, Seong~Joon Oh, Sangdoo Yun, Junsuk Choe, Eunji Kim, and Sungroh
  Yoon.
\newblock Weakly supervised semantic segmentation using out-of-distribution
  data.
\newblock In {\em CVPR}, pages 16897--16906, 2022.

\bibitem{lee2021bbam}
Jungbeom Lee, Jihun Yi, Chaehun Shin, and Sungroh Yoon.
\newblock Bbam: Bounding box attribution map for weakly supervised semantic and
  instance segmentation.
\newblock In {\em CVPR}, pages 2643--2652, 2021.

\bibitem{lee2021railroad}
Seungho Lee, Minhyun Lee, Jongwuk Lee, and Hyunjung Shim.
\newblock Railroad is not a train: Saliency as pseudo-pixel supervision for
  weakly supervised semantic segmentation.
\newblock In {\em CVPR}, pages 5495--5505, 2021.

\bibitem{lin2016scribblesup}
Di Lin, Jifeng Dai, Jiaya Jia, Kaiming He, and Jian Sun.
\newblock Scribblesup: Scribble-supervised convolutional networks for semantic
  segmentation.
\newblock In {\em CVPR}, pages 3159--3167, 2016.

\bibitem{long2015fully}
Jonathan Long, Evan Shelhamer, and Trevor Darrell.
\newblock Fully convolutional networks for semantic segmentation.
\newblock In {\em CVPR}, pages 3431--3440, 2015.

\bibitem{rossetti2022max}
Simone Rossetti, Damiano Zappia, Marta Sanzari, Marco Schaerf, and Fiora Pirri.
\newblock Max pooling with vision transformers reconciles class and shape in
  weakly supervised semantic segmentation.
\newblock In {\em ECCV}, pages 446--463. Springer, 2022.

\bibitem{shimoda2019self}
Wataru Shimoda and Keiji Yanai.
\newblock Self-supervised difference detection for weakly-supervised semantic
  segmentation.
\newblock In {\em ICCV}, pages 5208--5217, 2019.

\bibitem{sun2020mining}
Guolei Sun, Wenguan Wang, Jifeng Dai, and Luc Van~Gool.
\newblock Mining cross-image semantics for weakly supervised semantic
  segmentation.
\newblock In {\em ECCV}, pages 347--365. Springer, 2020.

\bibitem{wang2020self}
Yude Wang, Jie Zhang, Meina Kan, Shiguang Shan, and Xilin Chen.
\newblock Self-supervised equivariant attention mechanism for weakly supervised
  semantic segmentation.
\newblock In {\em CVPR}, pages 12275--12284, 2020.

\bibitem{wei2017object}
Yunchao Wei, Jiashi Feng, Xiaodan Liang, Ming-Ming Cheng, Yao Zhao, and
  Shuicheng Yan.
\newblock Object region mining with adversarial erasing: A simple
  classification to semantic segmentation approach.
\newblock In {\em CVPR}, pages 1568--1576, 2017.

\bibitem{wu2021embedded}
Tong Wu, Junshi Huang, Guangyu Gao, Xiaoming Wei, Xiaolin Wei, Xuan Luo, and
  Chi~Harold Liu.
\newblock Embedded discriminative attention mechanism for weakly supervised
  semantic segmentation.
\newblock In {\em CVPR}, pages 16765--16774, 2021.

\bibitem{wu2019wider}
Zifeng Wu, Chunhua Shen, and Anton Van Den~Hengel.
\newblock Wider or deeper: Revisiting the resnet model for visual recognition.
\newblock {\em Pattern Recognition}, 90:119--133, 2019.

\bibitem{xu2021scribble}
Jingshan Xu, Chuanwei Zhou, Zhen Cui, Chunyan Xu, Yuge Huang, Pengcheng Shen,
  Shaoxin Li, and Jian Yang.
\newblock Scribble-supervised semantic segmentation inference.
\newblock In {\em ICCV}, pages 15354--15363, 2021.

\bibitem{xu2022multi}
Lian Xu, Wanli Ouyang, Mohammed Bennamoun, Farid Boussaid, and Dan Xu.
\newblock Multi-class token transformer for weakly supervised semantic
  segmentation.
\newblock In {\em CVPR}, pages 4310--4319, 2022.

\bibitem{yao2020saliency}
Qi Yao and Xiaojin Gong.
\newblock Saliency guided self-attention network for weakly and semi-supervised
  semantic segmentation.
\newblock {\em IEEE Access}, 8:14413--14423, 2020.

\bibitem{zeng2019joint}
Yu Zeng, Yunzhi Zhuge, Huchuan Lu, and Lihe Zhang.
\newblock Joint learning of saliency detection and weakly supervised semantic
  segmentation.
\newblock In {\em ICCV}, pages 7223--7233, 2019.

\bibitem{zhang2018adversarial}
Xiaolin Zhang, Yunchao Wei, Jiashi Feng, Yi Yang, and Thomas~S Huang.
\newblock Adversarial complementary learning for weakly supervised object
  localization.
\newblock In {\em CVPR}, pages 1325--1334, 2018.

\bibitem{zhou2016learning}
Bolei Zhou, Aditya Khosla, Agata Lapedriza, Aude Oliva, and Antonio Torralba.
\newblock Learning deep features for discriminative localization.
\newblock In {\em CVPR}, pages 2921--2929, 2016.

\bibitem{zhou2022regional}
Tianfei Zhou, Meijie Zhang, Fang Zhao, and Jianwu Li.
\newblock Regional semantic contrast and aggregation for weakly supervised
  semantic segmentation.
\newblock In {\em CVPR}, pages 4299--4309, 2022.

\end{thebibliography}
}

\end{document}